# Machine learning models for estimating counterfactuals in a single-arm inflammatory bowel disease study

Dan Liu[1,2], Fida K. Dankar[1], Jennifer C. deBruyn[3,4], Amanda Ricciuto[5,6], Anne M. Griffiths[5,6], Thomas D. Walters[6], Khaled El Emam[1,2]

[1]*Children's Hospital of Eastern Ontario Research Institute, Ottawa, Ontario, Canada*
[2]*School of Epidemiology and Public Health, University of Ottawa, Ottawa, Ontario, Canada*
[3]*Alberta Children's Hospital, Calgary, Alberta, Canada*
[4]*University of Calgary, Calgary, Alberta, Canada*
[5]*University of Toronto, Toronto, Ontario, Canada*
[6]*Hospital for Sick Children, Toronto, Ontario, Canada*

Corresponding Author:
Khaled El Emam
Children's Hospital of Eastern Ontario Research Institute
401 Smyth Road
Ottawa, Ontario K1H 8L1
Canada

kelemam@ehealthinformation.ca

# Abstract

**Background:** In randomised clinical trials, enrolling patients into the control arm is essential to assess the effect of a new treatment, yet also challenging due to the high expenses and substantial time required. Single-arm trials accelerate study timelines by reducing the number of patients that must be recruited for a concurrent control group. However, these designs require an alternative comparator to estimate treatment effects. One approach is to construct a virtual control arm using a machine learning (ML) model trained on external control data to predict the counterfactual outcomes of the treatment arm.

**Objectives:** Our aim in this study is to leverage virtual controls by developing and evaluating ML-based counterfactual outcome models trained on IFX-treated patients to predict 1-year steroid-free clinical remission (SFCR ) and a composite of C-reactive protein remission plus steroid-free clinical remission (CRP-SFCR) for ADA-treated pediatric Crohn's disease patients, and to compare the resulting IFX-versus-ADA treatment effect estimates with those obtained using propensity score matching.

**Methods:** Five ML models, with and without augmentation, were used to train counterfactual models on the observed IFX cohort data, with augmentation implemented by adding synthetically generated IFX records to the training set. The resulting baseline and augmented models were used to predict the counterfactual outcomes for the ADA arm patients. Odds ratio (OR) with 95% confidence interval (CI) was calculated to assess the effect of IFX compared to ADA. These results were further compared to the results obtained using propensity score matching (reference standard), the default method for these types of comparative studies.

**Results:** Data augmentation using synthetic data generative models improved the performance of almost all the ML models significantly, with relative AUC and ICI up to 10% and 39%, respectively. Light gradient boosting machine yields the best augmented OR closest to the propensity score matched reference, and all 95% CI results align with the conclusion from the reference study that no statistical difference in the primary and secondary outcomes has been observed between the patients treated with ADA or IFX.

**Conclusions:** Our study supports virtual controls as a viable and effective substitute for expensive, lengthy or unethical patient recruitment in an inflammatory bowel disease (IBD) trial. The developed gradient boosted prediction model can be used as a pretrained model to generate IFX counterfactual predictions in future studies, pending external validation and assessment of transportability.

# 1 Introduction

A treatment effect compares a patient's outcome under treatment to the outcome that the same patient would experience under control. Because each patient can be observed under only one treatment condition, one potential outcome is observed and the other is unobserved—the counterfactual. Causal inference therefore depends on how well we can approximate this missing counterfactual with an appropriate comparison group. Randomised controlled trials (RCTs) address this by randomly assigning patients to treatment and control arms, balancing both observed and unobserved covariates so that outcomes in the control arm provide an unbiased estimate of the treated group's counterfactual outcomes (and vice versa). As a result, RCTs are widely regarded as the most robust design for estimating treatment effects in clinical research [1]. However, adequately powered RCTs are resource intensive, requiring recruitment and follow-up of a full comparator arm and imposing substantial operational and financial demands [2].

In fact, studies on clinical trials that led to new drug approvals have a median cost of about $41,000 per participant, with wide variations across studies ($31,020–$82,362) [3,4]. More recent industry cost models suggest higher overall trial expenses across all phases, with per-participant costs exceeding $100,000 and remaining relatively consistent between phases [5]. Another comparison study reported that among 138 pivotal studies, controlled trials were approximately 3-4 times more expensive than uncontrolled trials, with a median cost of $17.7 million for uncontrolled trials, and $56.7 million and $67.6 million for trials with placebo or active-drug comparators, respectively [4]. The largest cost-driving factor is the number of participants that must be recruited to sustain sufficient power in the trials [3,6].

To address these limitations, single-arm trials have emerged as a more practical and ethical alternative. It refers to a trial design, where only a treatment arm is established, without a concurrently randomised control arm. The primary benefit of single-arm trials is its ability to reduce costs and limit the number of patients who must forgo potentially beneficial therapies in favor of placebos or the standard of care (SOC) [7–9]. It is well-suited and has been applied to a variety of disease areas in clinical research, such as rare, infectious, oncological, and hematological diseases, etc., where conducting controlled trials is exceptionally challenging [10–14]. Recent studies show that single-arm trials have been increasingly promoted as an innovative trial design in support of new drug regulatory approvals in malignant oncology and hematology by the U.S. Food and Drug Administration (FDA) [15,16]. Another study reported that nearly half of FDA-accelerated approvals in oncology between 1992 and 2020 were received based on single-arm trials [17].

However, without a concurrently randomised control arm, estimating treatment effects requires an alternative method for approximating the outcomes patients would have experienced under an alternative treatment (i.e. counterfactual outcomes). In single-arm studies, the most common strategy for approximating these counterfactuals is to use external controls. Carefully selected external records - drawn from historical controls or contemporaneous cohorts - are used to assemble a comparator group that resembles the treatment arm as closely as possible [8,18–20]. By aligning the distributions of key prognostic and baseline covariates, these external controls aim to approximate the balance achieved through randomization, most commonly via propensity score matching, in which patients in the external control group are matched to patients in the treatment arm with similar characteristics, helping to isolate the effects of treatment from differences in patient populations [21].

An increasingly promising but still developing alternative is counterfactual outcome modeling, also known as *virtual controls* [22]. In this approach, statistical or machine learning (ML) models trained on external control data are used to predict the expected outcomes for the treatment arm participants. In other words, these models generate individualized counterfactuals, estimating what would have happened to each treated patient had they received the comparator therapy instead. They require sufficient training data and inclusion of the prognostic variables that influence the outcome to generate reliable predictions.

Virtual controls are hypothesised to provide two advantages over external controls in single-arm settings [22]. First, they are generated using predictive models that incorporate individual-level characteristics, thereby achieving closer alignment with the treatment group than is typically feasible with external controls. Second, because they do not require cohort selection from external datasets, they can reduce selection bias. In empirical evaluations, these outcome-modeling approaches have been reported to yield less biased and more statistically efficient estimates of treatment effects than traditional methods in settings where external data are well-aligned with the single-arm population [23,24]. They have also been found to provide more reliable estimates under certain forms of model misspecification [25].

To date, work on virtual controls has largely followed two paths: (i) developing a new model specifically for a given trial, or (ii) using a pretrained outcome model. The first approach develops trial-specific virtual controls by training ML models on external data corresponding to the trial's prespecified control, then using these models to estimate counterfactual outcomes [22,25]. When external data are unavailable or incomplete, pretrained models provide a practical and efficient means for counterfactual estimation and can be implemented more rapidly. These outcome models are developed prior to the current study and trained on independent datasets of patients treated under the control regimen, allowing them to serve as portable virtual controls that can be applied across multiple studies. Such models should be evaluated for generalizability to the study population and, if necessary, undergo minor recalibration. Pretrained virtual control models have been demonstrated in applications to breast cancer [26], prostate cancer [27], and cardiovascular disease [28].

In this paper, we use a previously published comparison of two monoclonal antibodies targeting tumour necrosis factor-α (anti-TNF), adalimumab (ADA) versus infliximab (IFX), in pediatric Crohn's disease [29] as a case study to develop and evaluate a virtual-control framework. Our aim is to evaluate whether virtual controls derived from observed IFX outcomes can support estimation of ADA/IFX treatment effects in a way that would be usable in a single-arm trial context, treating this cohort as a testbed for the virtual controls approach.

The original study was a non-randomised, prospective, real-world, multicenter cohort from the Canadian Children Inflammatory Bowel Disease (IBD) Network (CIDsCaNN) comparing ADA and IFX as first anti-TNF therapy in pediatric luminal Crohn's disease, with 1:1 propensity-score matching (PSM) (147 ADA vs 147 IFX). Using PSM, it evaluated 1-year steroid-free clinical remission (SFCR) as the primary outcome and a composite of normal serum C-reactive-protein (CRP) plus SFCR (CRP-SFCR) as a secondary outcome, finding comparable effectiveness between agents on both outcomes.

Our objectives are to develop models for both outcomes under IFX therapy and evaluate their agreement with the PSM-based estimates, specifically:

1. Use ML methods to train outcome models that predict 1-year SFCR and CRP-SFCR under IFX therapy.
2. Apply these models to ADA-treated patients to generate individualized virtual IFX outcomes and emulate a single-arm ADA trial augmented with virtual controls.
3. Compare treatment effect estimates obtained using virtual controls with those from the original propensity score matched ADA–IFX comparison, treating the latter as a reference standard.

As a secondary objective, given the relatively small IFX sample size available for model training, we assessed whether augmenting the IFX dataset with synthetic records improves the predictive performance of the ML models. Multiple synthetic data generation methods were compared, following the approach described in a recently published study [30]. To quantify the contribution of augmentation, all models were trained and evaluated under two settings: (i) the baseline IFX dataset only and (ii) the augmented IFX dataset.

Following its development and validation in this study, the resulting IFX outcome models could, in principle, be reused as pretrained models in future studies that enroll only a treatment arm and cannot feasibly recruit concurrent IFX controls. Such use would, however, require additional external validation and careful assessment of the model's generalizability.

## 2 Related Work

Trial-specific virtual controls train machine learning (ML) models directly on external SOC data to estimate counterfactual outcomes for participants of a single-arm cohort. In these approaches, the external dataset is carefully selected to match the inclusion criteria and covariate distribution of the treatment cohort, ensuring close alignment. Several studies have applied this strategy. Chen et al. presented a proof-of-concept framework in which a model for outcomes under control is trained on external control data and then used to impute the counterfactual control outcomes of participants in a single-arm study [31]. Loiseau et al. [23] demonstrated that these outcome-modeling approaches can yield less biased and more statistically efficient estimates of treatment effects than traditional propensity-score methods, provided that the external data are well-aligned with the single-arm population. Similarly, proof-of-concept applications in oncology [22] have used models trained on historical patients to predict expected progression-free survival under SOC, with treatment effects estimated by comparing these predictions to observed outcomes in the treated cohort. In the context of spinal cord injury, Lukas et al. [32] evaluated virtual controls constructed from individualized predictions of neurological recovery using multiple statistical and deep-learning models trained on a large observational cohort of patients receiving SOC. They identified convolutional neural networks as the best-performing architecture, and trial-simulation experiments showed that the resulting virtual controls produced recovery estimates comparable to randomised controls. Haussmann et al. introduced a model that learns jointly from single-arm trial data and external control data to estimate counterfactual outcomes [33] . In contrast to most prior applications, the model is trained on both external and trial data simultaneously to identify the factors that affect outcomes across both treated and control patients. Once trained, the model is used to generate counterfactuals. The authors evaluated their approach using a published RCT in chronic kidney disease in type 2 diabetes and an external EHR cohort, and they reported lower treatment effect estimation error than the baseline methods. Fang and Zhong [25] introduced a targeted virtual control approach for single-arm trials with external controls, where they update the initial outcome model predictions using a treatment assignment model, that is, a model for the probability of receiving treatment given patient characteristics. They call their method doubly robust because valid estimation can still be obtained if

6/21

either the outcome model or the treatment assignment model is correctly specified. However, this approach cannot be applied in the pretrained case because it requires access to external control data at the time of analysis.

In settings where a validated pretrained outcome model is available, such a model can provide a practical and efficient approach to counterfactual estimation, requiring less analytical effort and reducing the workload for trial sponsors. These models serve as portable virtual control generators, applicable across multiple trials to predict outcomes under standard treatment. However, as with any pre-trained model, they must be calibrated and periodically updated to reflect shifts in clinical practice [22].

Several groups have demonstrated that validated, disease-specific models can function as virtual controls in single-arm trials. For example, using the CancerMath breast cancer model, investigators in Switchenko et al. predicted SOC survival for participants in a mature adjuvant chemotherapy trial and showed that, after calibration in a similar population, the virtual controls closely reproduced observed control-arm outcomes, thereby supporting the treatment benefits reported in the original study [26]. In prostate cancer, virtual controls have been generated from Kattan's post-prostatectomy nomogram by using its progression-free survival predictions under SOC to construct counterfactual outcomes for high-risk cohorts [27]. Reported results showed close agreement between predicted and observed outcomes and greater sensitivity for detecting treatment benefit compared with propensity score based modeling approaches. Similarly, the Seattle Heart Failure tool has been applied to patients receiving left ventricular assist devices to estimate how they might have fared under standard medical therapy [28]. The model accurately stratified patients by risk and predicted outcomes consistent with observed survival, suggesting that validated disease-specific pretrained models may provide credible virtual controls when randomised comparisons are not feasible.

Taken together, existing applications indicate that virtual control models are a promising approach to estimating treatment effects when concurrent control groups are difficult to obtain. Nonetheless, additional validation is required to confirm their performance across diseases and data sources. In this context, our study adds to the literature by demonstrating that a model trained only on the IFX arm of the IBD inception cohort can generate counterfactual predictions for the ADA treatment group.

Our outcome modeling design relies on IFX data only and does not use information specific to the ADA arm, reflecting how a pretrained virtual control model is constructed. The model can be updated as additional IFX data become available, gradually evolving into a more robust pretrained virtual control model for use in single-arm IBD trials.

The paper is divided as follows. Section 3 describes the dataset derived from the Canadian Children Inflammatory Bowel Disease Network and details the methodology, including model training, counterfactual estimation, and data-manipulation procedures. Section 4 presents the results, followed by Section 5, which discusses the findings and their implications for future IBD trials. Section 6 concludes the paper.

# 3 Methods

## 3.1 Dataset

We used data from the CIDsCaNN, a prospective national multicenter cohort study of children and adolescents diagnosed with inflammatory bowel disease (IBD). CIDsCaNN includes all 12 major pediatric IBD centres across Canada. Details of the Network have been published previously [34]. The current analysis was restricted to participants with CD initiating anti-TNF therapy (IFX or ADA). The cohort included 613 children and 29 baseline predictor variables, pre-specified a priori for their clinical relevance to the outcomes. Two 1-year clinical outcomes were considered: SFCR as the primary outcome and a combined endpoint of SFCR and CRP (CRP-SFCR) as the secondary outcome [29]. Summary statistics for all baseline predictors and outcomes are provided in Table 1 in the Appendix.

Patients in this cohort received either IFX or ADA as anti-TNF therapy, with 437 children initiating IFX and 176 initiating ADA [29]. For model development, all 437 IFX-treated patients were used to train prognostic models for 1-year IFX outcomes. As mentioned above, a propensity score-matched analysis was previously reported [29] comparing IFX to ADA within the CIDsCaNN cohort. In that matched analysis, 147 ADA-treated patients (of the total 176 in the CIDsCaNN cohort) were used. For treatment-effect estimation with virtual controls, we used these 147 ADA-treated patients from the propensity score matched sample, allowing direct comparison with the previously reported analysis. In this framework, the IFX-treated group represents the SOC response, and the matched ADA cohort serves as the population for which counterfactual (virtual-control) outcomes under IFX are generated.

## 3.2 Models

A large variety of machine learning models have been used in the literature for training outcome models, including logistic regression, random forests, and other tree-based methods [22,26,35], highlighting the importance of comparing potential models. We developed and compared multiple models using the IFX data, employing both conventional non-pretrained ML models and a novel foundational pretrained model that is specifically designed for structured data prediction. The non-pretrained models include light gradient boosting machine (LGBM), extreme gradient boosting (XGB), random forest (RF), and logistic regression (LR), and the pretrained model is the tabular prior-data fitted network (TabPFN). TabPFN is a state-of-the-art transformer-based model that is pretrained on millions of synthetic datasets and designed to handle small tabular data in classification and regression tasks. It has been shown to outperform several well-known ML models on small and medium-sized datasets [36].

Given the relatively small IFX data size for model training, we also evaluated whether augmenting the IFX data with synthetic records could enhance the predictive performance of the ML models. Building on a recently published study [30], we leveraged multiple generative artificial intelligence models to augment the IFX dataset across a range of synthetic sample sizes, allowing systematic evaluation of different augmentation levels. To quantify the contribution of data augmentation within our overall virtual modeling strategy, we evaluated all models under two settings: (i) using the baseline IFX data only (Figure 1) and (ii) using the augmented IFX dataset (Figure 2).

### 3.3 Baseline model

This section describes the modeling approach and counterfactual estimation procedure applied to each of the five models, highlighting any differences as needed.

**Modeling.** We used nested 5-fold cross-validation (CV) for model training and evaluation, a strategy that has been shown to provide nearly unbiased estimates of model performance [37–39]. In each of the five CV iterations, we train a model on four folds of the IFX data and use the remaining fold for testing. Within each fold, we adopted Bayesian optimization to choose the best hyperparameter values. Since approximately half of the variables in the IFX data contained missing values, with proportions ranging from 2.3% to 57.9% (average 11.2%), we handled missingness using multiple imputation performed separately in the IFX training and IFX testing sets to obtain complete datasets [40,41]. Performing imputation independently within each set prevents data leakage that could lead to biased and overly optimistic performance estimates [42].

Bayesian optimization for hyperparameter tuning was applied only to the non-pretrained models, using recommended search ranges for each algorithm [43–46]. Throughout the analysis, non-pretrained predictive models were developed in R, while the pretrained model TabPFN was implemented in Python. The metrics used to evaluate the different classification models were the area under the receiver operating characteristic curve (AUC) and integrated calibration index (ICI). AUC quantifies the ability of a model to discriminate between outcome classes. It is one of the few recommended metrics to evaluate prediction models and is robust under class imbalance [47–49]. ICI is a calibration measure that numerically assesses model calibration as the weighted average difference between the observed and predicted probabilities [50]. For each model, AUC and ICI were computed in each test fold, and the final baseline values were obtained by averaging across the five iterations. These averaged metrics were used to summarize and report prognostic model performance.

**Counterfactuals.** We used the full imputed IFX dataset to train the final baseline prognostic models and then applied these models to the ADA arm to create a virtual control group composed of the same patients as the treatment arm, thereby emulating their outcomes had they received IFX instead. The workflow for baseline modeling and counterfactual outcome estimation is presented in Figure 1.
quantify the contribution of data augmentation

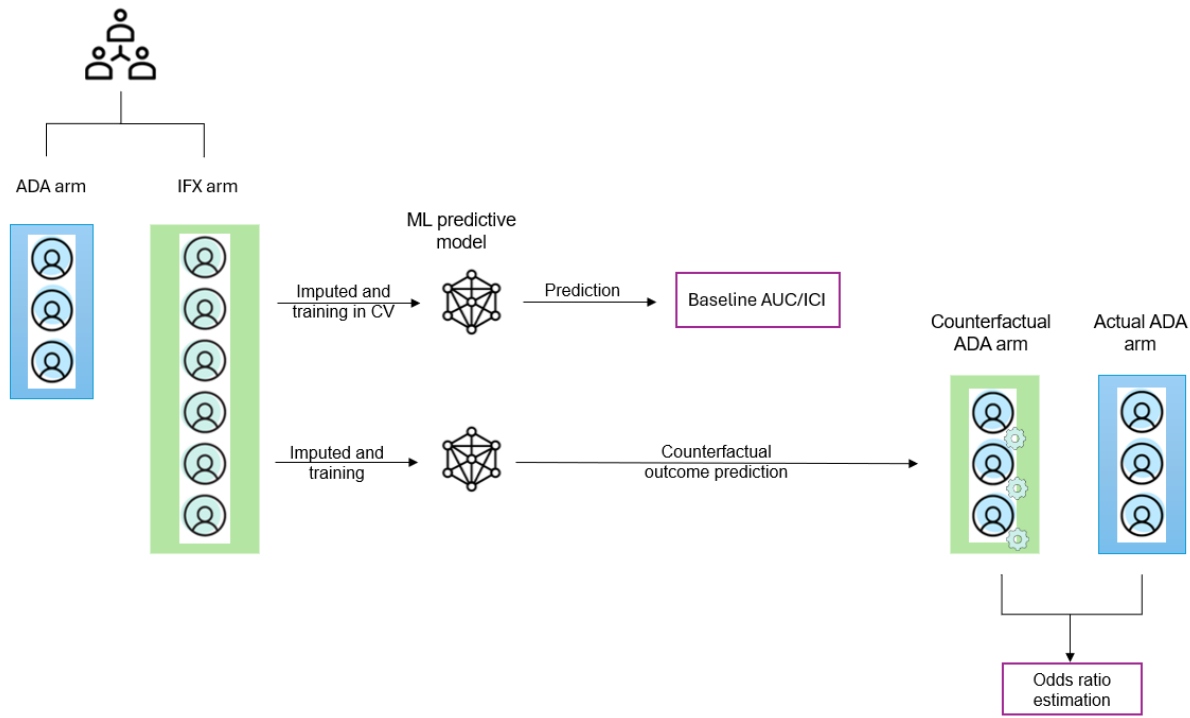

**Figure 1:** Workflow for baseline modeling and estimating the counterfactual outcomes.

Specifically, after completing the cross-validation procedure described in the prior section, we performed multiple imputation on the full IFX dataset, refit each prognostic model on the imputed IFX data using hyperparameters tuned via Bayesian optimization, and then used the refitted models to predict IFX counterfactual 1-year outcomes for the ADA-treated patients. This procedure yields a virtual IFX control arm that matches the ADA cohort on baseline characteristics.

We evaluated the comparative effectiveness of IFX versus ADA within the virtual controls framework by estimating odds ratios for each clinical outcome, comparing the counterfactual (virtual) and observed ADA arms. The 95% confidence interval (CI) was also computed using bootstrap resampling, with the significance level chosen to be 0.05. We compared these odds-ratio estimates with those from the propensity-score-matched analysis in [29]. The baseline results for both primary and secondary outcomes are reported in Table 1 and Table 2.

### 3.4 Augmented Model

**Augmentation Method.** To address the limited IFX sample size for model training, we used multiple synthetic data generation models to augment the IFX data with additional synthetic observations. A total of four common generative modeling methods were considered for the task: adversarial random forest (ARF) [51], Bayesian networks (BN) [52], conditional tabular generative adversarial network (CTGAN) [53] and tabular variational autoencoder (TVAE) [53]. The full description of each synthesis method is provided in Appendix. All models were implemented using pysdg, our adaptation of an open-sourced Python package Synthcity [54], which is publicly available, and provides further pre-processing and post-processing on top of Synthcity.

Following the method in [30], we considered a range of synthetic sample sizes that increase geometrically. Specifically, we drew 10 independent values of $b \sim N(\mu = 1.5, \sigma = 0.005)$, following a normal distribution with mean 1.5 and standard deviation 0.005. For each sampled value of $b$, we constructed a geometric series of 23 synthetic sample sizes as follows:

$$n_i' = [b^{i+4}], i = 1, \ldots, 23,$$

where $[\ \cdot\ ]$ denotes rounding to the closest integer.

Thus, using each generative model, we created synthetic samples for all the described sizes, yielding 230 augmented datasets per generative model. Each augmented dataset had a total training size $n = n_0 + n_i'$, where $n_0$ is the size of the original training data.

The augmented datasets were used to train the five ML models. To ensure comparability of results, within a given iteration, we evaluated all models trained on the different augmented datasets using the same test dataset.

**Choosing Augmentation Size.** Using the augmented datasets constructed above, we next aimed to select the degree of augmentation that best improves model performance. Within each generative-ML model combination, we used cross-validated AUC (with ICI as a secondary criterion) to identify the degree of augmentation that yielded the best predictive performance. The augmentation level selected in this step was then carried forward to construct the final augmented IFX dataset for counterfactual estimation. Figure 2 depicts the overall method to calculate the treatment effect for the data augmentation setting.

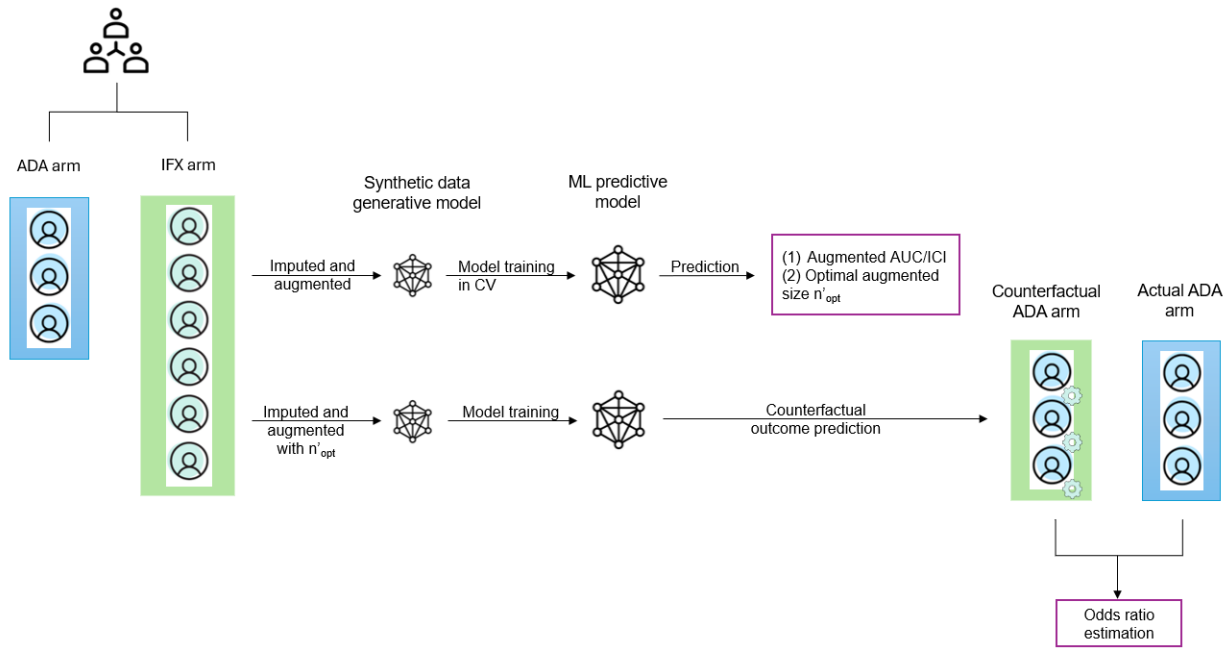

**Figure 2:** Workflow for augmented model performance and counterfactual estimation.

**Modeling and estimation.** The analysis of the augmented IFX data followed the same procedure as the baseline model, except that, in each iteration of the 5-fold cross-validation, the training data were first used by the generative models to simulate additional synthetic IFX records (according to the augmentation scheme described above). Subsequently, the original and synthetic records were combined to form an augmented training set for model fitting, while the corresponding test fold was left unchanged and used solely for performance evaluation.

The cross-validated model evaluation produced augmented AUC and ICI values for each combination of generative model and augmentation level, computed as the average across the 5 folds. Based on these values, we identified the optimal augmentation level for each generative model as the augmented sample size $n'_{opt}$ that maximized the augmented AUC, consistent with the procedure described above. Then, a synthetic dataset with the optimal sample size was generated and combined with the IFX data as the final augmented data for counterfactual model training and prediction.

Following the same estimation procedures as in the baseline model, a new ML model was built on the augmented IFX data, and the new counterfactual outcomes of the ADA patients were predicted from this constructed model under IFX. This provided a second set of virtual control outcomes, now based on models trained with augmented IFX data, enabling direct comparison of treatment effect estimates between the baseline and augmented analyses. The odds ratio and its 95% CI were estimated again to assess the comparative effectiveness, i.e. the association between the treatment and clinical outcome from the counterfactual and actual ADA arms. Table 1 and Table 2 summarize the baseline and augmentation results for both outcomes.

## 4     Results

Table 1 and Table 2 summarize the results for the primary and secondary outcomes, respectively. Each table reports (i) the predictive performance of the baseline and augmented models and (ii) the original and estimated treatment effects expressed as odds ratios (ORs) with 95% confidence intervals (CIs). The original ORs, treated as the ground truth in this comparative analysis, were extracted from [29] for the corresponding primary and secondary outcomes.

Across both tables, data augmentation generally enhances prognostic model performance, particularly for the primary outcome, as reflected by a higher augmented AUC and lower augmented ICI, resulting in up to approximately 10% discriminative performance gain and 39% calibration improvement. TabPFN achieves the highest baseline AUC and lowest baseline ICI in both outcome scenarios, demonstrating its ability to perform well on small datasets. However, no substantial improvements in augmented model performance are observed for TabPFN, which may be attributed to the sufficiently diverse large-scale synthetic datasets on which TabPFN was pretrained. As a result, further simulated data offer only limited marginal benefits for TabPFN, in contrast to the more pronounced improvements observed for the non-pretrained ML models on the IFX data.

In addition, for the primary outcome (SFCR), the closest agreement with the original treatment effect point estimate is achieved with LGBM combined with ARF (highlighted in bold), which yields an augmented OR closest to the original OR. Similarly, for the secondary outcome (CRP-SFCR), LGBM remains the only ML model that produces both baseline and augmented ORs with point estimates nearly equal to the original OR across all generative models (highlighted in bold). Although TabPFN also yields augmented ORs sufficiently close to the original values, LGBM provides OR estimates that are consistently closest to the original ORs across all settings. More importantly, the 95% CIs are broadly comparable across models and lead to the same conclusion as reported in [29] that no statistically significant difference has been detected between the ADA-treated and clinically similar IFX-treated patients for either outcome.

| Method | Model | Baseline AUC/ICI | Augmented AUC/ICI | Original OR | Baseline OR | Augmented OR |
| --- | --- | --- | --- | --- | --- | --- |
| **LGBM** | **ARF** | **0.62/0.10** | **0.67/0.08** | **1.4 (0.9, 2.4)** | **1.3 (0.5, 2.1)** | **1.4 (0.1, 2.6)** |
| LGBM | BN | 0.62/0.10 | 0.65/0.10 | 1.4 (0.9, 2.4) | 1.3 (0.5, 2.1) | 1.1 (0.4, 1.9) |
| LGBM | CTGAN | 0.62/0.10 | 0.65/0.10 | 1.4 (0.9, 2.4) | 1.3 (0.5, 2.1) | 1.3 (0.6, 2.0) |
| LGBM | TVAE | 0.62/0.10 | 0.66/0.08 | 1.4 (0.9, 2.4) | 1.3 (0.5, 2.1) | 1.1 (0.5, 1.7) |
| TabPFN | ARF | 0.65/0.07 | 0.67/0.10 | 1.4 (0.9, 2.4) | 0.9 (0.5, 1.2) | 1.0 (0.7, 1.3) |
| TabPFN | BN | 0.65/0.07 | 0.66/0.07 | 1.4 (0.9, 2.4) | 0.9 (0.5, 1.2) | 0.9 (0.5, 1.3) |
| TabPFN | CTGAN | 0.65/0.07 | 0.66/0.08 | 1.4 (0.9, 2.4) | 0.9 (0.5, 1.2) | 0.9 (0.5, 1.4) |
| TabPFN | TVAE | 0.65/0.07 | 0.66/0.08 | 1.4 (0.9, 2.4) | 0.9 (0.5, 1.2) | 0.9 (0.5, 1.3) |
| XGB | ARF | 0.62/0.11 | 0.66/0.07 | 1.4 (0.9, 2.4) | 1.1 (0.4, 1.8) | 1.1 (0.6, 1.7) |
| XGB | BN | 0.62/0.11 | 0.66/0.08 | 1.4 (0.9, 2.4) | 1.1 (0.4, 1.8) | 1.3 (0.6, 2.0) |
| XGB | CTGAN | 0.62/0.11 | 0.65/0.09 | 1.4 (0.9, 2.4) | 1.1 (0.4, 1.8) | 1.1 (0.3, 1.8) |
| XGB | TVAE | 0.62/0.11 | 0.64/0.09 | 1.4 (0.9, 2.4) | 1.1 (0.4, 1.8) | 1.1 (0.3, 1.8) |
| RF | ARF | 0.64/0.07 | 0.66/0.11 | 1.4 (0.9, 2.4) | 1.0 (0.4, 1.7) | 1.0 (0.5, 1.5) |
| RF | BN | 0.64/0.07 | 0.66/0.07 | 1.4 (0.9, 2.4) | 1.0 (0.4, 1.7) | 1.0 (0.5, 1.6) |
| RF | CTGAN | 0.64/0.07 | 0.66/0.08 | 1.4 (0.9, 2.4) | 1.0 (0.4, 1.7) | 0.9 (0.3, 1.5) |
| RF | TVAE | 0.64/0.07 | 0.66/0.07 | 1.4 (0.9, 2.4) | 1.0 (0.4, 1.7) | 1.0 (0.4, 1.5) |
| LR | ARF | 0.62/0.14 | 0.68/0.08 | 1.4 (0.9, 2.4) | 1.0 (0.4, 1.7) | 1.1 (0.7, 1.5) |
| LR | BN | 0.62/0.14 | 0.65/0.09 | 1.4 (0.9, 2.4) | 1.0 (0.4, 1.7) | 1.3 (0.7, 1.9) |
| LR | CTGAN | 0.62/0.14 | 0.64/0.12 | 1.4 (0.9, 2.4) | 1.0 (0.4, 1.7) | 1.0 (0.4, 1.6) |
| LR | TVAE | 0.62/0.14 | 0.64/0.12 | 1.4 (0.9, 2.4) | 1.0 (0.4, 1.7) | 1.0 (0.4, 1.5) |

**Table 1:** Model performance in the IFX arm and odds ratio estimation of ADA patients for SFCR outcome. AUC: area under the ROC curve. ICI: integrated calibration index. OR: odds ratio estimate. LGBM: light gradient boosting machine. TabPFN: tabular prior-data fitted network. XGB: extreme gradient boosting. RF: random forest. LR: logistic regression. ARF: adversarial random fores. BN: Bayesian networks. CTGAN: conditional tabular generative adversarial network. TVAE: tabular variational autoencoder.

| Method | Model | Baseline AUC/ICI | Augmented AUC/ICI | Original OR | Baseline OR | Augmented OR |
| --- | --- | --- | --- | --- | --- | --- |
| **LGBM** | **ARF** | **0.63/0.07** | **0.67/0.08** | **1.0 (0.6, 1.6)** | **1.0 (0.3, 1.8)** | **1.0 (0.1, 1.9)** |
| **LGBM** | **BN** | **0.63/0.07** | **0.65/0.08** | **1.0 (0.6, 1.6)** | **1.0 (0.3, 1.8)** | **1.0 (0.3, 1.7)** |
| **LGBM** | **CTGAN** | **0.63/0.07** | **0.65/0.09** | **1.0 (0.6, 1.6)** | **1.0 (0.3, 1.8)** | **1.0 (0.2, 1.7)** |
| **LGBM** | **TVAE** | **0.63/0.07** | **0.65/0.09** | **1.0 (0.6, 1.6)** | **1.0 (0.3, 1.8)** | **1.0 (0.4, 1.6)** |
| TabPFN | ARF | 0.67/0.07 | 0.67/0.09 | 1.0 (0.6, 1.6) | 0.9 (0.5, 1.3) | 1.0 (0.7, 1.4) |
| TabPFN | BN | 0.67/0.07 | 0.67/0.06 | 1.0 (0.6, 1.6) | 0.9 (0.5, 1.3) | 1.0 (0.6, 1.4) |
| TabPFN | CTGAN | 0.67/0.07 | 0.67/0.07 | 1.0 (0.6, 1.6) | 0.9 (0.5, 1.3) | 1.0 (0.6, 1.4) |
| TabPFN | TVAE | 0.67/0.07 | 0.67/0.08 | 1.0 (0.6, 1.6) | 0.9 (0.5, 1.3) | 1.0 (0.6, 1.4) |
| XGB | ARF | 0.61/0.11 | 0.66/0.09 | 1.0 (0.6, 1.6) | 1.1 (0.4, 1.7) | 1.2 (0.6, 1.9) |
| XGB | BN | 0.61/0.11 | 0.64/0.11 | 1.0 (0.6, 1.6) | 1.1 (0.4, 1.7) | 1.1 (0.4, 1.7) |
| XGB | CTGAN | 0.61/0.11 | 0.64/0.09 | 1.0 (0.6, 1.6) | 1.1 (0.4, 1.7) | 1.1 (0.6, 1.7) |
| XGB | TVAE | 0.61/0.11 | 0.65/0.08 | 1.0 (0.6, 1.6) | 1.1 (0.4, 1.7) | 1.1 (0.5, 1.8) |
| RF | ARF | 0.62/0.08 | 0.65/0.09 | 1.0 (0.6, 1.6) | 1.0 (0.3, 1.7) | 1.1 (0.4, 1.7) |
| RF | BN | 0.62/0.08 | 0.63/0.09 | 1.0 (0.6, 1.6) | 1.0 (0.3, 1.7) | 1.1 (0.4, 1.7) |
| RF | CTGAN | 0.62/0.08 | 0.64/0.08 | 1.0 (0.6, 1.6) | 1.0 (0.3, 1.7) | 0.9 (0.4, 1.4) |
| RF | TVAE | 0.62/0.08 | 0.64/0.08 | 1.0 (0.6, 1.6) | 1.0 (0.3, 1.7) | 1.1 (0.5, 1.7) |
| LR | ARF | 0.64/0.11 | 0.69/0.10 | 1.0 (0.6, 1.6) | 1.1 (0.4, 1.8) | 1.1 (0.7, 1.5) |
| LR | BN | 0.64/0.11 | 0.65/0.10 | 1.0 (0.6, 1.6) | 1.1 (0.4, 1.8) | 1.0 (0.5, 1.6) |
| LR | CTGAN | 0.64/0.11 | 0.66/0.11 | 1.0 (0.6, 1.6) | 1.1 (0.4, 1.8) | 1.2 (0.6, 1.8) |
| LR | TVAE | 0.64/0.11 | 0.66/0.08 | 1.0 (0.6, 1.6) | 1.1 (0.4, 1.8) | 1.0 (0.4, 1.6) |

**Table 2:** Model performance in the IFX arm and odds ratio estimation of ADA patients for CRP-SFCR outcome. AUC: area under the ROC curve. ICI: integrated calibration index. OR: odds ratio estimate. LGBM: light gradient boosting machine. TabPFN: tabular prior-data fitted network. XGB: extreme gradient boosting. RF: random forest. LR: logistic regression. ARF: adversarial random fores. BN: Bayesian networks. CTGAN: conditional tabular generative adversarial network. TVAE: tabular variational autoencoder.

# 5 Discussion

## 5.1 Summary

In randomised clinical trials, it is often challenging to conduct studies with both treatment and control arms because of limited budgets, slow recruitment, and ethical concerns about assigning patients to non-experimental care. This has motivated the development of single-arm trial designs enriched with external or virtual control arms as alternative approaches.

In this article, we evaluated the virtual controls approach in estimating treatment effects in a prospective non-randomised IBD trial. The central idea was to use the trial's control arm as the SOC reference and to train a machine learning model to generate virtual control outcomes (i.e., simulated patient outcomes under SOC) to serve as the comparator for the observed treatment group. The analysis used the inception-cohort Crohn's disease data from a prospective multicenter cohort study. Unmatched infliximab-treated patients from this cohort were treated as the control data. We considered five ML models: four conventional, non-pretrained approaches (random forest, LightGBM, XGBoost, and logistic regression) and one pretrained model, the Tabular Prior-Data Fitted Network (TabPFN). The first four are widely used for clinical prediction, whereas TabPFN is a recently introduced transformer-based foundation model designed for small- to medium-sized tabular datasets.

We adopted a two-stage analysis strategy. The initial analysis was to train all these models on the IFX data to get the baseline model performance and to predict the counterfactual outcomes of patients in the ADA arm. The second part of the analysis augmented the existing small IFX data using one of the four synthetic data generative models, before training and using the model to recalculate the counterfactual outcomes.

Our estimation results agree with the findings in [29] that from a statistical viewpoint, no evidence has been found in support of the assertion that IFX-treated patients have superior outcomes to ADA-treated patients. We also find that none of the models show evidence that IFX leads to better outcomes than ADA, as the estimated odds ratios are consistently close to 1 and their confidence intervals include 1 across both outcomes. When we apply synthetic data augmentation, the predictive performance generally improves, with higher AUC and lower ICI for most model-generator combinations, achieving discrimination and calibration improvements of up to 10% and 39%, respectively. In addition, the resultant AUC and ICI values suggest that the current LGBM model is possibly better at calibration than discrimination, yielding predicted probabilities sufficiently close to the true ones. At the same time, augmentation makes the no-superiority conclusion clearer, with estimated odds ratios tending to move closer to 1 and confidence intervals becoming more centered around 1.

Across both outcomes, augmented LGBM performs best overall, achieving the strongest combination of high discrimination (AUC) and good calibration (low ICI). TabPFN shows comparatively smaller gains from augmentation, consistent with its extensive pretraining and reduced sensitivity to additional synthetic samples.

## 5.2 Recommendations for Research

The main practical contribution of this study is to provide a proof of concept that virtual controls can be constructed from an existing control cohort and used for comparative effectiveness assessment in IBD. In our case, we had access to a concurrent IFX control arm and used it to train models that predict counterfactual outcomes for ADA-treated patients. This setting allows us to test the approach and illustrate how it could be used in future trials where a concurrent control arm is not available. In such situations, similar models could be trained on external control data.

In practice, external data should be selected to closely match the target trial in terms of eligibility criteria, baseline characteristics, follow-up, outcome definitions, and SOC, ensuring a credible comparator for counterfactual estimation. Potential sources include prior trial arms, registries, and electronic health records, provided they capture a comparable patient trajectory. Precedents and methodological guidance on selecting appropriate external data for a given trial are already available and can be leveraged when constructing virtual controls in IBD [55–57]. The LGBM model developed in this study could serve as a starting point for a pretrained model that is further updated with additional data in future IBD trials.

## 5.3 Limitations and Future Work

To move from this proof of concept toward a more robust pretrained model for future IBD trials, the current LGBM model will need to be updated and expanded with additional data to improve the accuracy and transportability of its counterfactual predictions. Another IBD study is currently underway with more patients enrolled into the IFX cohort; we plan to validate the present model on these new data and then update it accordingly. In the longer term, incorporating data from external cohorts will be important to test whether a pretrained virtual-control model can be adapted across different study populations and settings.

In our comparison, we relied on a non-randomised inception study in which IFX and ADA were observed concurrently, and the analysis used propensity score matching. Although matching is common in observational work, it is not equivalent to randomization, so our findings should be interpreted as agreement with a strong non-randomised comparison rather than validation against a randomised trial. Residual confounding and selection bias therefore cannot be completely excluded, and the estimated treatment effects should be read with this limitation in mind. The IFX and ADA data were collected with the same selection criteria at approximately the same time, at the same sites, and under the same treatment regimen, which strengthens internal comparability. However, this also means that model performance should be evaluated in other cohorts, across different centres, healthcare systems, and treatment regimens to assess generalizability and to guard against overfitting to the present dataset.

## Ethics

This project was approved by the CHEO Research Institute Research Ethics Board, protocol number: CHEOREB:# 26/50X.

## Funding Statement

This research is funded by the Canadian Children Inflammatory Bowel Disease Network, the Canada Research Chairs program through the Canadian Institutes of Health Research, and a Discovery Grant RGPIN-2022-04811 from the Natural Sciences and Engineering Research Council of Canada.

## Code Availability

The code used in this analysis can be accessed as follows:

- The synthetic data generation code is available in the pysdg package, available from: <https://osf.io/xj9pr/>

- The machine learning modeling was performed using the R sdgm package available from: <https://osf.io/DCJM6>

# Appendix

# Machine learning models for estimating counterfactuals in a single-arm inflammatory bowel disease study

## Contents

# Appendix A – Inception cohort data

| Variable | Description | Type | Mean (SD) or level count (% of total size) or number of categories | Missingness (% of the total size) |
|---|---|---|---|---|
| SFCR | Steroid-free clinical remission (primary outcome) | Categorical | 1: 56.3%<br>0: 38.0% | 5.7 |
| CRP-SCFR | SCFR plus CRP remission (secondary outcome | Categorical | 1: 49.6%<br>0: 40.5% | 10.0 |
| Agent | Type of anti-TNF agents (either Infliximab or Adalimumab) | Categorical | 1: 28.7%<br>0: 71.3% | 0.0 |
| Age | Patient's age in months | Numeric | 24.4 (5.3) | 0.0 |
| Sex | Patient's gender | Categorical | 1: 60.7%<br>0: 39.3% | 0.0 |
| L1-3, ideally L4 | Disease classification of L1-3 at diagnosis | Categorical | 0: 2.8%<br>1: 16.8%<br>2: 23.5%<br>3: 54.6% | 2.3 |
| B1-B3 | Disease classification of B1-B3 at diagnosis | Categorical | 1: 89.6%<br>2: 7.5%<br>3: 1.1%<br>4: 1.8% | 0.0 |
| Perianal involvement | Disease classification of perianal involvement at diagnosis | Categorical | 1: 17.0%<br>0: 82.9% | 0.2 |
| Weighted pediatric Crohn's disease activity index | Clinical activity at diagnosis | Numeric | 9.5 (4.2) | 5.9 |
| CRP | Biochemical activity of CRP at diagnosis | Numeric | 7.3 (7.8) | 11.1 |
| Albumin | Biochemical activity of albumin at diagnosis | Numeric | 5.4 (1.1) | 10.1 |
| Haemoglobin | Biochemical activity of haemoglobin at diagnosis | Numeric | 17.9 (2.6) | 7.7 |
| ESR | Biochemical activity of ESR at diagnosis | Numeric | 6.8 (4.2) | 21.2 |
| Endoscopic activity | Biochemical activity of endoscopic activity at diagnosis | Numeric | 2.7 (1.4) | 23.7 |
| Disease duration | Disease duration | Numeric | 1.0 (1.4) | 0.0 |
| Weighted pediatric Crohn's disease activity index | Clinical activity at anti-TNF start | Numeric | 7.0 (4.7) | 0.5 |
| CRP | Biochemical activity of CRP at anti-TNF start | Numeric | 5.2 (6.9) | 4.1 |
| Albumin | Biochemical activity of albumin at anti-TNF start | Numeric | 5.8 (1.3) | 4.1 |
| Haemoglobin | Biochemical activity of | Numeric | 18.5 (2.8) | 2.8 |

| | haemoglobin at anti-TNF start | | | |
|---|---|---|---|---|
| ESR | Biochemical activity of ESR at anti-TNF start | Numeric | 5.3 (4.0) | 11.6 |
| Faecal calprotectin | Biochemical activity of faecal calprotectin at anti-TNF start | Numeric | 310.7 (373.4) | 57.9 |
| Exclusive enteral nutrition | Treatment history for exclusive enteral nutrition | Categorical | 1: 25.1% 2: 15.2% 3: 59.7% | 0.0 |
| Prior steroids exposure | Treatment history for steroids prior to anti-TNF start | Categorical | 1: 19.7% 2: 25.0% 3: 55.3% | 0.0 |
| Prior immunomodulator monotherapy trial (azathioprine) | Treatment history for prior immunomodulator monotherapy trial - azathioprine | Categorical | 1: 17.9% 0: 82.1% | 0.0 |
| Prior immunomodulator monotherapy trial (methotrexate) | Treatment history for prior immunomodulator monotherapy trial - methotrexate | Categorical | 1: 29.2% 0: 70.8% | 0.0 |
| Steroids | Treatment history for steroids at time of starting anti-TNF | Categorical | 1: 25.0% 0: 75.0% | 0.0 |
| Exclusive enteral nutrition at time of starting anti-TNF | Treatment history for exclusive enteral nutrition at time of starting anti-TNF | Categorical | 1: 15.2% 0: 84.8% | 0.0 |
| Combo immunomodulator use | Treatment history for combo immunomodulator use with anti-TNF | Categorical | 1: 56.0% 0: 44.0% | 0.0 |
| Combo immunomodulator use (azathioprine) | Treatment history for combo immunomodulator use with anti-TNF - azathioprine | Categorical | 1: 13.5% 0: 86.5% | 0.0 |
| Combo immunomodulator use (methotrexate) | Treatment history for combo immunomodulator use with anti-TNF - methotrexate | Categorical | 1: 42.4% 0: 57.6% | 0.0 |
| Concomitant immunomodulator use | Whether concomitant immunomodulator is used | Categorical | 1: 43.6% 0: 56.4% | 0.0 |

**Table 1:** A summary of statistics descriptive of the variables in the unmatched inception cohort data.

# Appendix B – Hyperparameters

The following are the hyperparameters, their default values, and the range for tuning the ensemble models.

| LGBM | | | | |
|---|---|---|---|---|
| **Hyperparameter** | **Default** | **Lower bound** | **Upper bound** | **Transform** |
| booster | 1 (gbdt) | 1 (gbdt) | 2 (goss) | 2^learning_rate |
| max_depth | 6 | 1 | 15 | |
| learning_rate | log2 (0.3) | -10 | 0 | |
| early_stopping_rounds | 7 | 7 | 30 | |
| min_data_in_leaf | 10 | 1 | 60 | |
| num_leaves | 15 | 4 | 60 | |
| **Random forest** | | | | |
| **Hyperparameter** | **Default** | **Lower bound** | **Upper bound** | **Transform** |
| num.trees | 500 | 1 | 2000 | round(n^min.node.size), where n is the number of observations |
| min.node.size | 0.5 | 0 | 1 | |
| max.depth | 15 | 1 | 50 | |
| min.bucket | 10 | 1 | 60 | |
| **XGBoost** | | | | |
| **Hyperparameter** | **Default** | **Lower bound** | **Upper bound** | **Transform** |
| gamma | 0 | -15 | 3 | 2^gamma |
| eta | log2(0.3) | -10 | 0 | 2^eta |
| max_depth | 6 | 1 | 15 | |
| early_stopping_rounds | 7 | 7 | 30 | |
| max_leaves | 15 | 4 | 60 | |
| min_child_weight | 1 | 0 | 7 | 2^min_child_weight |

**Table 2:** The default values and ranges of hyperparameters in the models.

# Appendix C – Synthetic data generative models

**Adversarial random forests**

Adversarial Random Forests is a tree-based density estimator that uses recursive unsupervised random forests [1]. Inspired by GANs, ARFs employ a recursive process where trees iteratively learn the structural properties of data by alternating between rounds of data generation and discrimination. This allows the model to gradually refine its understanding of the data distribution. Unlike classic tree-based models, ARFs provide smooth density estimations and can generate fully synthetic data.

**Bayesian networks**

Bayesian Networks (BN) are models based on Directed Acyclic Graphs that consist of nodes representing the random variables and arcs representing the dependencies among these variables. To construct the BN model, the first step is to find the optimal network topology, and then to estimate the optimal parameters [2]. Starting with a random initial network structure, the Hill Climb heuristic search is used to find the optimal structure. Then, the conditional probability distributions are estimated using the maximum a posteriori estimator [3]. Once the network structure and the parameters are estimated, we can initialize the nodes with no incoming arcs by sampling from their marginal distributions and predict the rest of the connected variables using the estimated parameters.

**Conditional tabular generative adversarial networks**

A basic generative adversarial network (GAN) consists of two artificial neural networks (ANNs), a generator and a discriminator [4]. The generator and the discriminator play a min-max game. The input to the generator is noise, while its output is synthetic data. The discriminator has two inputs: the real training data and the synthetic data generated by the generator. The output of the discriminator indicates whether its input is real or synthetic. The generator is trained to 'trick' the discriminator by generating samples that look real. On the other hand, the discriminator is trained to maximize its discriminatory capability.

Among all the variations of GAN architectures, the conditional tabular GAN (CTGAN) is often used in tabular data synthesis [5]. CTGAN builds on conditional GANs by addressing the multimodal distributions of continuous variables and the highly imbalanced categorical variables [6]. CTGAN solves the first problem by proposing a per-mode normalization technique. For the second problem, each category of a categorical variable serves as the condition passed to the GAN.

**Tabular variational autoencoder**

Variational autoencoders (VAE) use ANNs and involve two steps (encoding and decoding) to generate new samples [7]. First, an encoder is generated to compress input data into a lower-dimensional latent space, in which the data points are represented by distributions. The second step is a decoding process, in which new data samples are reconstructed as output from the latent space. The neural network is optimized by minimizing the reconstruction loss between the output and the input. VAEs are known to generate complex data of various types due to its ability to learn more complex distributions [8]. Many variants have been proposed as an extension of VAE, such as triplet-based VAE [9], conditional VAE [10], and Gaussian VAE [11]. In particular, the tabular VAE (TVAE) was proposed as an adaption of standard VAE to model and generate mixed-type tabular data with a modified loss function [6].